\documentclass{article}

\usepackage{times}
\usepackage{graphicx} % more modern
\usepackage{subfigure} 

\usepackage{natbib}

\usepackage{algorithm}
\usepackage{algorithmic}

\usepackage{hyperref}

\usepackage[accepted]{icml2017}

\icmltitlerunning{Averaged-DQN: Variance Reduction and Stabilization for Deep Reinforcement Learning}

\usepackage{amsmath,amssymb,amsthm}
\usepackage{dsfont}
\usepackage{setspace}

\newcommand{\Qstr}{Q^*(s,a)}
\newcommand{\Qi}{Q(s,a;\theta_i)} 

\newcommand{\AQiunderMax}{Q^{A}_{i-1}(s',a')}
\newcommand{\AQiExplicit}{\frac{1}{K}\sum^{K}_{k=1} Q(s,a;\theta_{i-k})}
\newcommand{\AQi}{Q^{A}_{i-1}(s,a)}
\newcommand{\AveragedDQNOutput}{Q^{A}_{N}(s,a) = \frac{1}{K}\sum^{K-1}_{k=0} Q(s,a;\theta_{N-k})}

\newcommand{\EQi}{Q^{E}_{i-1}(s,a)}
\newcommand{\EQiExplicit}{\frac{1}{K}\sum^{K}_{k=1} Q(s,a;\theta^{k}_{i-1})}
\newcommand{\EQiunderMax}{Q^{E}_{i-1}(s',a')}
\newcommand{\yiEnsembleExplict}{\mathbb{E}_{\mathcal{B}}\left[r + \gamma \max_{a'} \EQiunderMax)|\,s,a\right]}
\newcommand{\EnsembleDQNOutput}{Q^{E}_{N}(s,a) = \frac{1}{K}\sum^{K}_{k=1} Q(s,a;\theta^k_{i})}

\newcommand{\yi}{y^i_{s,a}}
\newcommand{\yiHat}{ \hat{y}^i_{s,a}}

\newcommand{\yiExplict}{\mathbb{E}_{\mathcal{B}}\left[r + \gamma \max_{a'} Q(s',a';\theta_{i-1})|\,s,a\right]}
\newcommand{\yiHatExplict}{\mathbb{E}_{\mathcal{B}}\left[r + \gamma \max_{a'} (y^{i-1}_{s',a'})|\,s,a\right]}
\newcommand{\yiAveragedExplict}{\mathbb{E}_{\mathcal{B}}\left[r + \gamma \max_{a'} \AQiunderMax|\,s,a\right]}

\newcommand{\DQNloss}{\mathbb{E}_{\mathcal{B}} \left[ (\yi-Q(s,a;\theta))^2  \right]}

\newcommand{\ExperienceTransition}{(s,a,r,s')}

\newcommand{\ApproxErr}{{target approximation error}}
\newcommand{\OverErr}{{overestimation error}}

\newcommand{\OverVar}{R^{i}_{s,a}}
\newcommand{\ApproxVar}{Z^{i}_{s,a}}

\newcommand{\Algo}[1]{Algorithm {#1}}
\newcommand{\Figure}[1]{Figure {#1}}

\begin{document} 

\twocolumn[
\icmltitle{Averaged-DQN: Variance Reduction and Stabilization \\ for Deep Reinforcement Learning}



% you can specify symbols, otherwise they are numbered in order
\icmlsetsymbol{equal}{*}

\begin{icmlauthorlist}
\icmlauthor{Oron Anschel}{teh}
\icmlauthor{Nir Baram}{teh}
\icmlauthor{Nahum Shimkin}{teh}
\end{icmlauthorlist}

\icmlaffiliation{teh}{Department of Electrical Engineering, Haifa 32000, Israel}

\icmlcorrespondingauthor{Oron Anschel}{oronanschel@campus.technion.ac.il}
\icmlcorrespondingauthor{Nir Baram}{nirb@campus.technion.ac.il}
\icmlcorrespondingauthor{Nahum Shimkin}{shimkin@ee.technion.ac.il}

\icmlkeywords{Reinforcement Learning, Deep Learning, Variance Reduction, DQN, Function Approximation, Arcade Learning Environment}
\vskip 0.3in
]



\printAffiliationsAndNotice{}  % leave blank if no need to mention equal contribution
%\footnotetext{hi}

\begin{abstract}
Instability and variability of Deep Reinforcement Learning (DRL) algorithms tend to adversely affect their performance.
Averaged-DQN is a simple extension to the DQN algorithm, based on averaging previously learned Q-values estimates, which leads to a more stable training procedure and improved performance by reducing approximation error variance in the target values.
To understand the effect of the algorithm, we examine the source of value function estimation errors and provide an analytical comparison within a simplified model.
We further present experiments on the Arcade Learning Environment benchmark that demonstrate significantly improved stability and performance due to the proposed extension.
\end{abstract}

\section{Introduction}
% RL problem
In Reinforcement Learning (RL) an agent seeks an optimal policy for a sequential decision making problem \citep{sutton1998reinforcement}.
It does so by learning which action is optimal for each environment state.
% RL algorithms
Over the course of time, many algorithms have been introduced for solving RL problems including Q-learning \citep{watkins1992q}, SARSA \citep{rummery1994line,sutton1998reinforcement}, and policy gradient methods \citep{sutton1999policy}.
These methods are often analyzed in the setup of linear function approximation, where convergence is guaranteed under mild assumptions \citep{tsitsiklis1994asynchronous, jaakkola1994convergence,tsitsiklis1997analysis,even2003learning}.
% Linear function approx
In practice, real-world problems usually involve high-dimensional inputs forcing linear function approximation methods to rely upon hand engineered features for problem-specific state representation.
% Problem of Linear function approx
These problem-specific features diminish the agent flexibility, and so the need of an expressive and flexible non-linear function approximation emerges.
Except for few successful attempts (e.g., TD-gammon, \citet{tesauro1995temporal}), the combination of non-linear function approximation and RL was considered unstable and was shown to diverge even in simple domains \citep{boyan1995generalization}.\par 
% DQN and
The recent Deep Q-Network (DQN) algorithm \citep{mnih2013playing}, was the first to successfully combine a powerful non-linear function approximation technique known as Deep Neural Network (DNN) \citep{lecun1998gradient,krizhevsky2012imagenet} together with the Q-learning algorithm.
DQN presented a remarkably flexible and stable algorithm, showing success in the majority of games within the Arcade Learning Environment (ALE) \citep{bellemare13arcade}.
DQN increased the training stability by breaking the RL problem into sequential supervised learning tasks.
To do so, DQN introduces the concept of a target network and uses an Experience Replay buffer (ER) \citep{lin1993reinforcement}.

Following the DQN work, additional modifications and extensions to the basic algorithm further increased training stability.
\citet{schaul2015prioritized} suggested sophisticated ER sampling strategy.
Several works extended standard RL exploration techniques to deal with high-dimensional input \citep{bellemare2016unifying,countExploration,osband2016deep}.
\citet{mnih2016asynchronous} showed that sampling from ER could be replaced with asynchronous updates from parallel environments (which enables the use of on-policy methods).
\citet{dueling} suggested a network architecture base on the advantage function decomposition \citep{baird1993advantage}.

In this work we address issues that arise from the combination of Q-learning and function approximation.
\citet{thrun1993issues} were first to investigate one of these issues which they have termed as the \emph{overestimation phenomena}.
The max operator in Q-learning can lead to overestimation of state-action values in the presence of noise.
\citet{van2015deep} suggest the Double-DQN that uses the Double Q-learning estimator \citep{double_q_learning} method as a solution to the problem.
Additionally, \citet{van2015deep} showed that Q-learning overestimation do occur in practice (at least in the ALE).

This work suggests a different solution to the overestimation phenomena, named Averaged-DQN (Section \ref{sec:ADQN}), based on averaging previously learned Q-values estimates.
The averaging reduces the target approximation error variance (Sections \ref{sec:oversetimation_and_approx} and \ref{sec:variance_reduction}) which leads to stability and improved results.
Additionally, we provide experimental results on selected games of the Arcade Learning Environment. 

We summarize the main contributions of this paper as follows:
\begin{itemize}
\item A novel extension to the DQN algorithm which stabilizes training, and improves the attained performance, by averaging over previously learned Q-values.
\item Variance analysis that explains some of the DQN problems, and how the proposed extension addresses them.
\item Experiments with several ALE games demonstrating the favorable effect of the proposed scheme. 
\end{itemize}

\section{Background}
In this section we elaborate on relevant RL background, and specifically on the Q-learning algorithm.
\subsection{Reinforcement Learning}
We consider the usual RL learning framework \citep{sutton1998reinforcement}.
An agent is faced with a sequential decision making problem, where interaction with the environment takes place at discrete time steps ($t=0,1,\ldots$). At time $t$ the agent observes state $s_t \in S$, selects an action $a_t \in A$, which results in a scalar reward $r_t \in \mathbb{R}$, and a transition to a next state $s_{t+1}\in S$.
We consider infinite horizon problems with a discounted cumulative reward objective $R_t =\sum^{\infty}_{t'=t} \gamma^{t'-t} r_{t'}$, where $\gamma \in [0,1]$ is the discount factor. The goal of the agent is to find an optimal policy $\pi: S \rightarrow A $ that maximize its expected discounted cumulative reward.

Value-based methods for solving RL problems encode policies through the use of value functions, which denote the expected discounted cumulative reward from a given state $s$, following a policy $\pi$. Specifically we are interested in state-action value functions:
\begin{equation*}
Q^{\pi}(s,a)= \mathbb{E}^{\pi}\left[\sum^{\infty}_{t=0} \gamma^{t} r_t \,\,| s_0=s , a_0 = a \right].
\end{equation*}
The optimal value function is denoted as $Q^{*}(s,a)= \max_\pi Q^{\pi}(s,a) $, and an optimal policy $\pi^*$ can be easily derived by $\pi^*(s) \in \text{argmax}_a Q^{*}(s,a)$.

\subsection{Q-learning}
One of the most popular RL algorithms is the Q-learning algorithm \citep{watkins1992q}.
This algorithm is based on a simple value iteration update \citep{Bel}, directly estimating the optimal value function $Q^{*}$.
Tabular Q-learning assumes a table that contains old action-value function estimates and preform updates using the following update rule:
\begin{equation}
\label{eq:q_learning_update_rule}
Q(s,a) \leftarrow Q(s,a)+\alpha(r+\gamma \max_{a'}Q(s',a')-Q(s,a)),
\end{equation}
where $s'$ is the resulting state after applying action $a$ in the state $s$, $r$ is the immediate reward observed for action $a$ at state $s$, $\gamma$ is the discount factor, and $\alpha$ is a learning rate.
\par
When the number of states is large, maintaining a look-up table with all possible state-action pairs values in memory is impractical.
A common solution to this issue is to use function approximation parametrized by $\theta$, such that $Q(s,a)\approx Q(s,a;\theta)$.

\subsection{Deep Q Networks (DQN)}
\label{sec:DQN}
We present in Algorithm~\ref{algo:DQN} a slightly different formulation of the DQN algorithm \citep{mnih2013playing}.
In iteration $i$ the DQN algorithm solves a supervised learning problem to approximate the action-value function $Q(s,a;\theta)$ (line 6).
This is an extension of implementing  \eqref{eq:q_learning_update_rule} in its function approximation form \citep{riedmiller2005neural}. 

\begin{algorithm}
\caption{DQN} \label{algo:DQN}
\begin{spacing}{1.2}
\begin{algorithmic}[1]
\STATE Initialize $Q(s,a;\theta)$ with random weights $\theta_0$
\STATE Initialize Experience Replay (ER) buffer $\mathcal{B}$
\STATE Initialize exploration procedure \emph{Explore($\cdot$)}
\FOR{$i=1,2,\ldots,N$}
    \STATE $ \yi =\yiExplict$
    \STATE $\theta_i \approx \text{argmin}_\theta \,\, \DQNloss$
    \STATE \emph{Explore($\cdot$)}, update $\mathcal{B}$
\ENDFOR
\OUTPUT $Q^{\text{DQN}}(s,a;\theta_N)$
\end{algorithmic}
\end{spacing}
\end{algorithm}

The target values $\yi$ (line 5) are constructed using a designated \emph{target-network} $Q(s,a;\theta_{i-1})$ (using the previous iteration parameters $\theta_{i-1}$), where the expectation ($\mathbb{E}_\mathcal{B}$) is taken w.r.t.\ the sample distribution of experience transitions  in the ER buffer $\ExperienceTransition \sim \mathcal{B}$.
The DQN loss (line 6) is minimized  using a Stochastic Gradient Descent (SGD) variant, sampling mini-batches from the ER buffer.
Additionally, DQN requires an exploration procedure (which we denote as \emph{Explore($\cdot$)}) to interact with the environment (e.g., an $\epsilon$-greedy exploration procedure).
The number of new experience transitions $\ExperienceTransition$ added  by exploration to the ER buffer in each iteration is small, relatively to the size of the ER buffer.
Thereby, $\theta_{i-1}$ can be used as a good initialization for $\theta$ in iteration $i$.

Note that in the original implementation \citep{mnih2013playing,mnih2015human}, transitions are added to the ER buffer simultaneously with the minimization of the DQN loss (line 6).
Using the hyperparameters employed by \citet{mnih2013playing,mnih2015human} (detailed for completeness in Appendix~\ref{appendix:dqn_hyper_params}), 1\% of the experience transitions in ER buffer are replaced between target network parameter updates, and 8\% are sampled for minimization.

\section{Averaged DQN}
\label{sec:ADQN}
\begin{figure}[ht]
\vskip 0.2in
\begin{center}
\centerline{\includegraphics[width=\columnwidth]{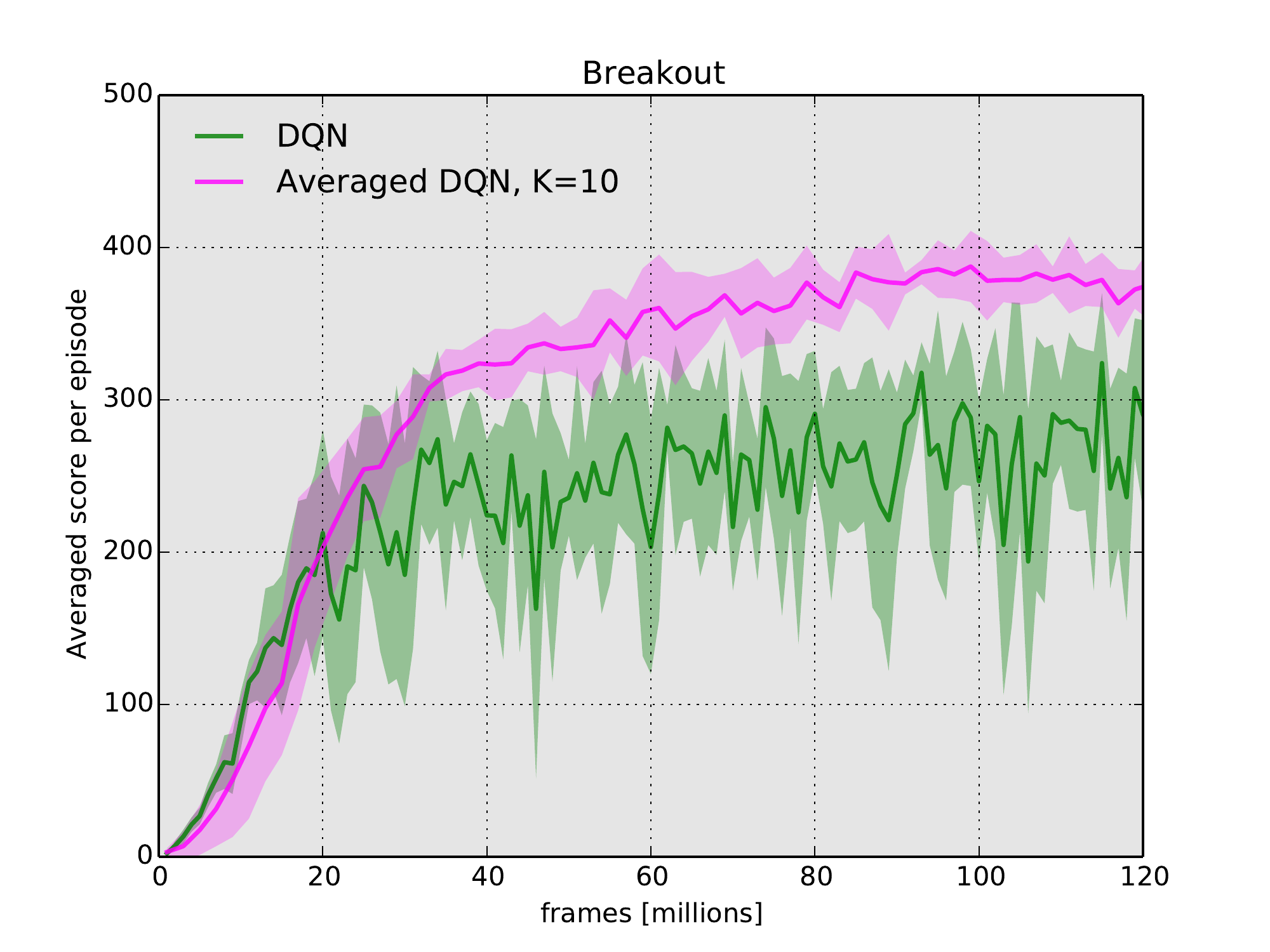}}
\caption{DQN and Averaged-DQN performance in the Atari game of \textsc{Breakout}.
The bold lines are averages over seven independent learning trials.
Every 1M frames, a performance test using $\epsilon$-greedy policy with $\epsilon=0.05$ for 500000 frames was conducted.
The shaded area presents one standard deviation.
For both DQN and Averaged-DQN the hyperparameters used were taken from \citet{mnih2015human}.
}
\label{fig:Approx and Over Err Example}
\end{center}
\vskip -0.2in
\end{figure}

The Averaged-DQN algorithm (\Algo{\ref{algo:ADQN}}) is an extension of the DQN algorithm.
Averaged-DQN uses the $K$ previously learned Q-values estimates to produce the current action-value estimate (line 5).
The Averaged-DQN algorithm stabilizes the training process (see \Figure{\ref{fig:Approx and Over Err Example}}), by reducing the variance of \emph{target approximation error} as we elaborate in Section \ref{sec:variance_reduction}.
The computational effort compared to DQN is, $K$-fold more forward passes through a Q-network while minimizing the DQN loss (line 7).
The number of back-propagation updates (which is the most demanding computational element), remains the same as in DQN.
The output of the algorithm is the average over the last $K$ previously learned Q-networks.

\begin{algorithm}
\caption{Averaged DQN} \label{algo:ADQN}
\begin{spacing}{1.2}
\begin{algorithmic}[1]
\STATE Initialize $Q(s,a;\theta)$ with random weights $\theta_0$
\STATE Initialize Experience Replay (ER) buffer $\mathcal{B}$
\STATE Initialize exploration procedure \emph{Explore}$(\cdot)$
\FOR{$i=1,2,\ldots,N$}
	\STATE $ \AQi = \AQiExplicit $
	\STATE $ \yi =\yiAveragedExplict$
	\STATE $\theta_i \approx \text{argmin}_\theta \,\, \DQNloss$
	\STATE \emph{Explore($\cdot$)}, update $\mathcal{B}$
\ENDFOR
\OUTPUT $\AveragedDQNOutput$
\end{algorithmic}
	\end{spacing}
\end{algorithm}

\begin{figure*}[ht]
\vskip 0.2in
\begin{center}
\centerline{
\includegraphics[width=0.85\columnwidth]{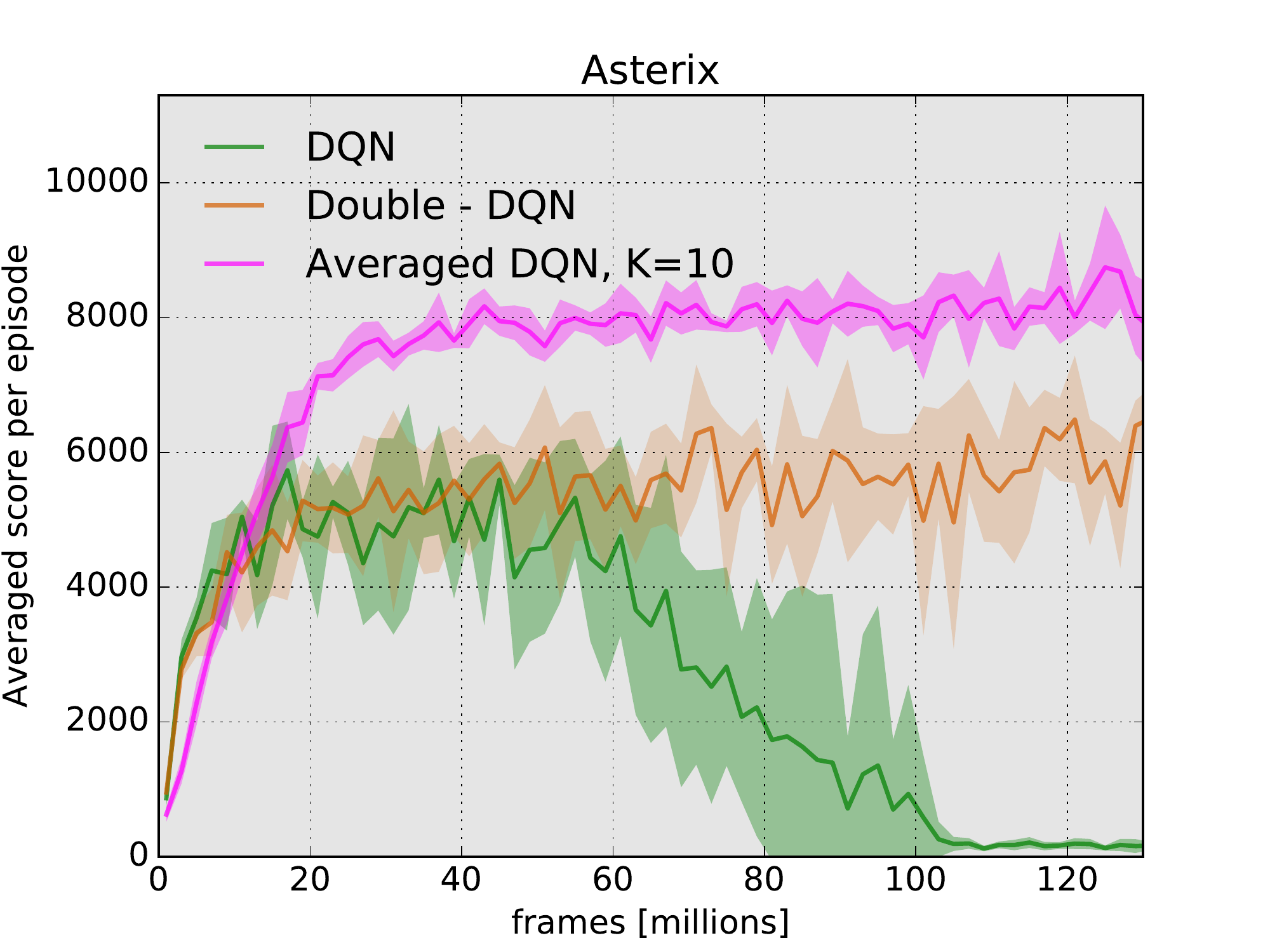}
\includegraphics[width=0.85\columnwidth]{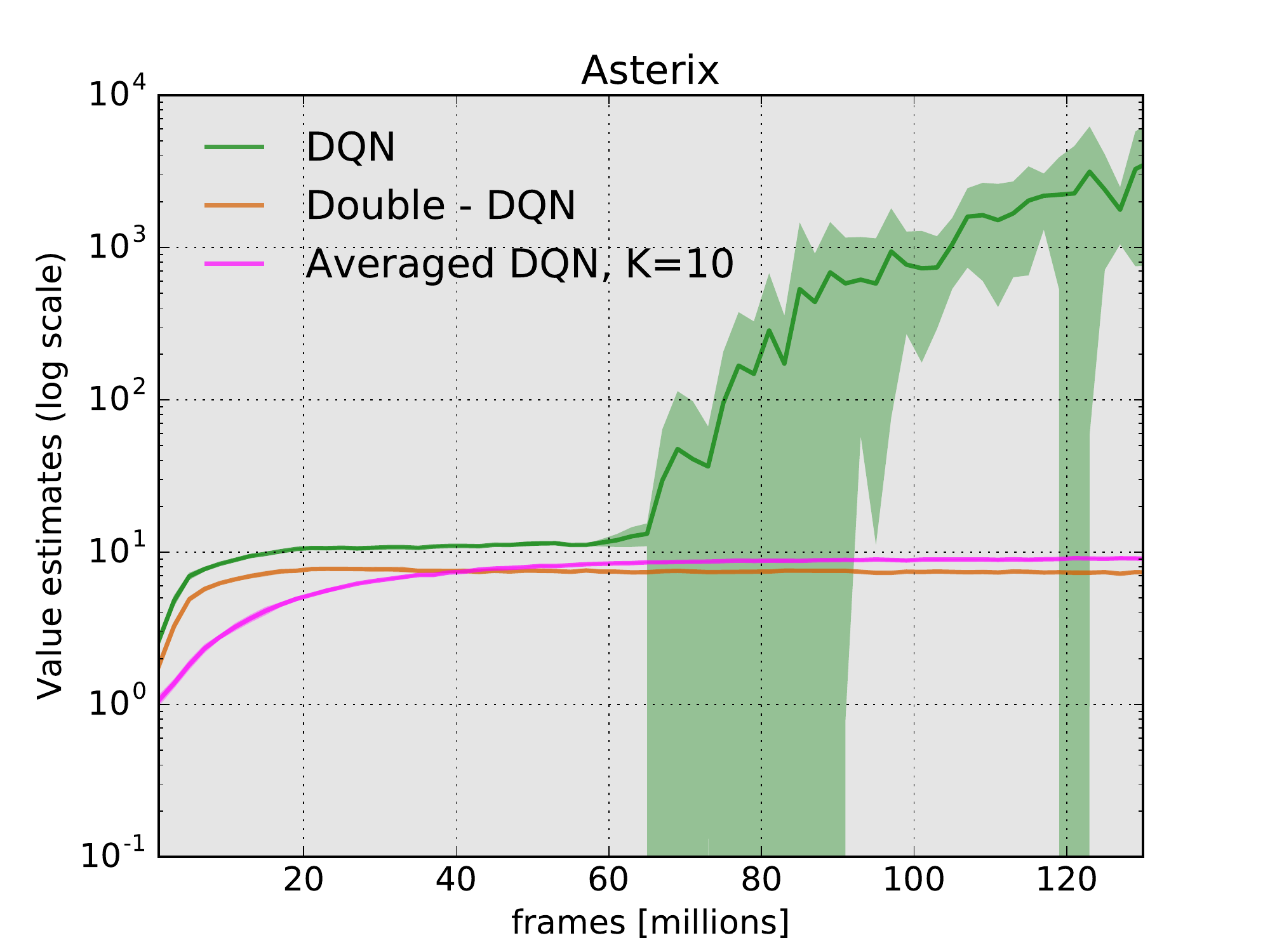}
}
\caption{DQN, Double-DQN, and Averaged-DQN performance (left), and average value estimates (right) in the Atari game of \textsc{Asterix}.
The bold lines are averages over seven independent learning trials.
The shaded area presents one standard deviation.
Every 2M frames, a performance test using $\epsilon$-greedy policy with $\epsilon=0.05$ for 500000 frames was conducted.
The hyperparameters used were taken from \citet{mnih2015human}.
}
\label{fig:asymototic_sub_optimal_asterix}
\end{center}
\vskip -0.2in
\end{figure*}

In Figures~\ref{fig:Approx and Over Err Example} and \ref{fig:asymototic_sub_optimal_asterix} we can see the performance of Averaged-DQN compared to DQN (and Double-DQN), further experimental results are given in Section~\ref{sec:experiments}. 

We note that recently-learned state-action value estimates are likely to be better than older ones, therefore we have also considered a recency-weighted average.
In practice, a weighted average scheme did not improve performance and therefore is not presented here.

\section{Overestimation and Approximation Errors}
\label{sec:oversetimation_and_approx}
Next, we discuss the various types of errors that arise due to the combination of Q-learning and function approximation in the DQN algorithm, and their effect on training stability.
We refer to DQN's performance in the \textsc{Breakout} game in \Figure {\ref{fig:Approx and Over Err Example}}.
The source of the learning curve variance in DQN's performance is an occasional sudden drop in the average score that is usually recovered in the next evaluation phase (for another illustration of the variance source see Appendix~\ref{appendix:variance_between_models}). 
Another phenomenon can be observed in \Figure{\ref{fig:asymototic_sub_optimal_asterix}}, where DQN initially reaches a steady state (after 20 million frames), followed  by a gradual deterioration in performance.

For the rest of this section, we list the above mentioned errors, and discuss our hypothesis as to the relations between each error and the instability phenomena depicted in Figures \ref{fig:Approx and Over Err Example} and \ref{fig:asymototic_sub_optimal_asterix}.

We follow terminology from \citet{thrun1993issues}, and define some additional relevant quantities.
Letting $\Qi$ be the value function of DQN at iteration $i$, we denote $\Delta_i = \Qi-\Qstr$ and decompose it as follows:

\begin{align*}
\Delta_i &= \Qi-\Qstr\\
  &=  \underbrace{\Qi-\yi}_{\substack{\text{\emph{Target Approximation}} \\ \text{\emph{Error}}}}  +
 \underbrace{\yi-\yiHat}_{\substack{\text{\emph{Overestimation}} \\ \text{\emph{Error}}}} + \underbrace{\yiHat - \Qstr}_{\substack{\text{\emph{Optimality}} \\ \text{\emph{Difference}}}}.
\end{align*}
Here $\yi$ is the \emph{DQN target}, and $\yiHat$ is the \emph{true target}:
\begin{align*}
\yi &=\yiExplict, \\
\yiHat &= \yiHatExplict.
\end{align*}

Let us denote by $\ApproxVar$ the {\ApproxErr}, and by $\OverVar$ the {\OverErr}, namely
\begin{align*}
\ApproxVar &= \Qi - \yi,\\
\OverVar &=\yi-\yiHat.
\end{align*}

The optimality difference can be seen as the error of a standard tabular Q-learning, here we address the other errors.
We next discuss each error in turn.

\subsection{Target Approximation Error (TAE)}
The TAE ($\ApproxVar$), is the error in the learned $\Qi$ relative to $\yi$, which is determined after minimizing the DQN loss (\Algo{\ref{algo:DQN}} line 6, \Algo{\ref{algo:ADQN}} line 7).
The TAE is a result of several factors:
Firstly, the sub-optimality of $\theta_i$ due to inexact minimization.
Secondly, the limited representation power of a neural net (model error).
Lastly, the generalization error for unseen state-action pairs due to the finite size of the ER buffer. 

\par
The TAE can cause a deviations from a policy to a worse one.
For example, such deviation to a sub-optimal policy occurs in case $\yi = \yiHat = \Qstr$ and,
\begin{align*}
\text{argmax}_a [\Qi ] &\neq \text{argmax}_a [\Qi - \ApproxVar ]\\
&=\text{argmax}_a [\yi ].
\end{align*}

We hypothesize that the variability in DQN's performance in \Figure{\ref{fig:Approx and Over Err Example}}, that was discussed at the start of this section, is related to deviating from a steady-state policy induced by the TAE.

\subsection{Overestimation Error}
The Q-learning \emph{overestimation phenomena} were first investigated by \citet{thrun1993issues}.
In their work, Thrun and Schwartz considered the TAE $\ApproxVar$ as a random variable uniformly distributed in the interval $[-\epsilon,\epsilon]$.
Due to the $\max$ operator in the {DQN target} $\yi$, the expected \OverErr s $\mathbb{E}_z[\OverVar]$ are upper bounded by $\gamma\epsilon\frac{n-1}{n+1}$ (where $n$ is the number of applicable actions in state $s$).
The intuition for this upper bound is that in the worst case, all $Q$ values are equal, and we get equality to the upper bound:
\begin{align*}
\mathbb{E}_z [\OverVar] = \gamma \mathbb{E}_z [\max_{a'}[Z^{i-1}_{s',a'}] ]= \gamma\epsilon\frac{n-1}{n+1}.
\end{align*}

The \OverErr\, is different in its nature from the TAE since it presents a positive bias that can cause asymptotically sub-optimal policies, as was shown by \citet{thrun1993issues}, and later by \citet{van2015deep} in the ALE environment.
Note that a uniform bias in the action-value function will not cause a change in the induced policy. Unfortunately, the overestimation bias is uneven and is bigger in states where the Q-values are similar for the different actions, or in states which are the start of a long trajectory (as we discuss in Section~\ref{sec:variance_reduction} on accumulation of TAE variance).
\par 
Following from the above mentioned overestimation upper bound, the magnitude of the bias is controlled by the variance of the TAE.

The Double Q-learning and its DQN implementation (Double-DQN) \cite{van2015deep,double_q_learning} is one possible approach to tackle the overestimation problem, which replaces the positive bias with a negative one. Another possible remedy to the adverse effects of this error is to directly reduce the variance of the TAE, as in our proposed scheme (Section \ref{sec:variance_reduction}).

In \Figure{\ref{fig:asymototic_sub_optimal_asterix}} we repeated the experiment presented in \citet{van2015deep} (along with the application of Averaged-DQN).
This experiment is discussed in \citet{van2015deep} as an example of overestimation that leads to asymptotically sub-optimal policies.  
Since Averaged-DQN reduces the TAE variance, this experiment supports an hypothesis that the main cause for overestimation in DQN is the TAE variance.

%An example where overestimation leads to asymptotically sub-optimal policies, that was discussed and presented in \citet{van2015deep}, in the game \textsc{Asterix} can be observed in \Figure{\ref{fig:asymototic_sub_optimal_asterix}}. Our hypothesis regarding this case study is that after achieving a steady state in performance, the overestimation bias is gradually learned, and therefore the performance declines.

%\subsection{Optimality Difference}
%We decompose the difference between the target value $\yi$ and the optimal value $\Qstr$, to the sum of two errors. The overestimation error $\OverVar$, and the error induced from not constructing the target value $\yi$ based on the previous target value $y^{i-1}_{s,a}$ (as if we had a look-up table implementation of DQN), and using its approximated version instead $Q(s,a;\theta_{i-1})$.

%
%The optimality difference is the error resulted from not using directly the previous target value for the current one, and using a value which is based on a learned approximation instead.

%
%In case DQN is implemented using a look-up table, the TAE and overestimation due to the TAE does not exist.
%Under mild assumptions, which mainly require the exploration to perform each action in each state infinitely often, the optimality difference converges to zero with probability one \citep{watkins1992q,jaakkola1994convergence,szepesvari1997asymptotic,even2003learning}.

\section{TAE Variance Reduction}
\label{sec:variance_reduction}
To analyse the TAE variance we first must assume a statistical model on the TAE, and we do so in a similar way to \citet{thrun1993issues}.
Suppose that the TAE $\ApproxVar$ is a random process such that $\mathbb{E}[\ApproxVar]=0$, $\text{Var}[\ApproxVar] = \sigma^2_s $, and for $i \neq j$: $\text{Cov}[\ApproxVar,Z^{j}_{s',a'}]=0$.
Furthermore, to focus only on the TAE we eliminate the overestimation error by considering a fixed policy for updating the target values.
Also, we can conveniently consider a zero reward $r=0$ everywhere since it has no effect on variance calculations.

\begin{figure}[ht]
\vskip 0.2in
\begin{center}
\centerline{
\includegraphics[width=1\columnwidth]{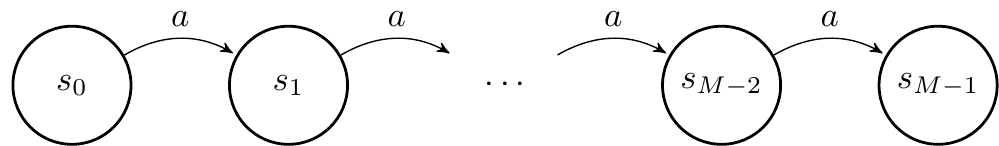}
}
\caption{$M$ states unidirectional MDP, The process starts at state $s_0$, then in each time step moves to the right, until the terminal state $s_{M-1}$ is reached. A zero reward is obtained in any state.
}
\label{fig:one_directional_chain_mdp}
\end{center}
\vskip -0.2in
\end{figure}

Denote by $Q_i \triangleq Q(s;\theta_i)_{s\in S}$ the vector of value estimates in iteration $i$ (where the fixed action $a$ is suppressed), and by $Z_i$ the vector of corresponding TAEs.
For Averaged-DQN we get:
\begin{align*}
Q_i = Z_i + \gamma P\frac{1}{K}\sum^{K}_{k=1} Q_{i-k},
\end{align*}

where $P\in \mathbb{R}_{+}^{S\times S}$ is the transition probabilities matrix for the given policy.
Assuming stationarity of $Q_i$, its covariance can be obtained using standard techniques (e.g., as a solution of a linear equations system).
However, to obtain an explicit comparison, we further specialize the model to an $M$-state unidirectional MDP as in Figure~\ref{fig:one_directional_chain_mdp}

\subsection{DQN Variance} 
We assume the statistical model mentioned at the start of this section.
Consider a unidirectional Markov Decision Process (MDP) as in \Figure{\ref{fig:one_directional_chain_mdp}}, where the agent starts at state $s_0$, state $s_{M-1}$ is a terminal state, and the reward in any state is equal to zero.

Employing DQN on this MDP model, we get that for $i>M$:
\begin{align*}
\label{eq:variance_dqn_unidirectional_mdp}
& Q^{\text{DQN}}(s_0,a;\theta_i) =  Z^{i}_{s_0,a} + y^{i}_{s_0,a} \\
& = Z^{i}_{s_0,a} + \gamma Q(s_1,a;\theta_{i-1}) \\
& = Z^{i}_{s_0,a} + \gamma [Z^{i-1}_{s_1,a} + y^{i-1}_{s_1,a}] = \cdots = \\
& = Z^{i}_{s_0,a} + \gamma Z^{i-1}_{s_1,a} + \cdots +  \gamma^{(M-1)} Z^{i-(M-1)}_{s_{M-1},a},\\
\end{align*}
where in the last equality we have used the fact $y^{j}_{M-1,a}=0$ for all $j$ (terminal state).
Therefore,
\begin{align*}
 \text{Var}[Q^{\text{DQN}}(s_0,a;\theta_{i})] = \sum_{m=0}^{M-1} \gamma^{2m} \sigma^2_{s_m}.
\end{align*}

The above example gives intuition about the behavior of the TAE variance in DQN.
The TAE is accumulated over the past DQN iterations on the updates trajectory.
Accumulation of TAE errors results in bigger variance with its associated adverse effect, as was discussed in Section~\ref{sec:oversetimation_and_approx}.

\begin{algorithm}
\caption{Ensemble DQN} \label{algo:Ensemble_DQN}
\begin{spacing}{1.2}
\begin{algorithmic}[1]
\STATE Initialize $K$ Q-networks $Q(s,a;\theta^k)$ with random weights $\theta^k_0$ for $k
\in \{1,\ldots,K\}$
\STATE Initialize Experience Replay (ER) buffer $\mathcal{B}$
\STATE Initialize exploration procedure \emph{Explore($\cdot$)}
\FOR{$i=1,2,\ldots,N$}
	\STATE $ \EQi = \EQiExplicit $
	\STATE $ \yi =\yiEnsembleExplict$
	\FOR{$k=1,2,\ldots,K$}
		\STATE $\theta^{k}_i \approx \text{argmin}_\theta \,\, \DQNloss$
	\ENDFOR  
	\STATE \emph{Explore($\cdot$)}, update $\mathcal{B}$
\ENDFOR
\OUTPUT $\EnsembleDQNOutput$
\end{algorithmic}
\end{spacing}
\end{algorithm}

\subsection{Ensemble DQN Variance}
\label{sec:Ensemble DQN Variance}
We consider two approaches for TAE variance reduction.
The first one is the Averaged-DQN and the second we term \emph{Ensemble-DQN}.
We start with Ensemble-DQN which is a straightforward way to obtain a $1/K$ variance reduction, with a computational effort of $K$-fold learning problems, compared to DQN.
Ensemble-DQN (Algorithm \ref{algo:Ensemble_DQN}) solves $K$ DQN losses in parallel, then averages
over the resulted Q-values estimates.

For Ensemble-DQN on the unidirectional MDP in Figure \ref{fig:one_directional_chain_mdp}, we get for $i>M$:
\begin{align*}
 Q^{E}_i(s_0,a) &= \sum^{M-1}_{m=0} \gamma^m  \frac{1}{K}\sum^{K}_{k=1} Z^{k,i-m}_{s_m,a}, \\
 \text{Var}[Q^E_i(s_0,a)] &=     \sum_{m=0}^{M-1} \frac{1}{K} \gamma^{2m} \sigma^2_{s_m}\\
  &=  \frac{1}{K} \text{Var}[Q^{\text{DQN}}(s_0,a;\theta_{i})],
\end{align*}
where for $k\neq k'$: $Z^{k,i}_{s,a}$ and $Z^{k',j}_{s',a'}$ are two uncorrelated TAEs.
The calculations of $Q^E(s_0,a)$ are detailed in Appendix \ref{appendix:ensemble_dqn_two_state_cal}.

\subsection{Averaged DQN Variance}
\label{sec:Averaged DQN Variance}
We continue with Averaged-DQN, and calculate the variance in state $s_0$ for the unidirectional MDP in Figure \ref{fig:one_directional_chain_mdp}. We get that for $i>KM$:
%\begin{align*}
%Q^{A}_i(s_0,a) &= \frac{1}{K}\sum^{K}_{k=1}Z^{i+1-k}_{s_0,a} +\\
%&+ \frac{\gamma}{K^2}\sum^{K}_{k=1}\sum^{K}_{k'=1} Z^{i+1-k-k'}_{s_{1},a}\\
%&\vdots\\
%&+ \frac{\gamma^{M-1}}{K^M}\sum^{K}_{j_1=1}\sum^{K}_{j_2=1}\cdots \sum^{K}_{j_{M}=1} Z^{i+1-j_1-\cdots -j_M}_{s_{M-1},a},\\
%\end{align*}
\begin{align*}
\text{Var}[Q^{A}_i(s_0,a)] &= \sum_{m=0}^{M-1} D_{K,m} \gamma^{2m} \sigma^2_{s_m},
\end{align*}
where $D_{K,m}=\frac{1}{N} \sum^{N-1}_{n=0}|U_n/K|^{2(m+1)}$, with $U=(U_n)_{n=0}^{N-1}$ denoting a Discrete Fourier Transform (DFT) of a rectangle pulse, and $|U_n/K|\leq 1$.
The calculations of $Q^A(s_0,a)$ and $D_{K,m}$ are more involved and are detailed in Appendix \ref{appendix:adqn_two_state_cal}.

Furthermore, for $K>1,m>0$ we have that $D_{K,m}< 1/K$ (Appendix~\ref{appendix:adqn_two_state_cal}) and therefore the following holds
\begin{align*}
\text{Var}[Q^{A}_i(s_0,a)] &<  \text{Var}[Q^{E}_i(s_0,a)] \\
&=  \frac{1}{K} \text{Var}[Q^{\text{DQN}}(s_0,a;\theta_{i})],
\end{align*}

meaning that Averaged-DQN is theoretically more efficient in TAE variance reduction than Ensemble-DQN, and at least $K$ times better than DQN.
The intuition here is that Averaged-DQN averages over TAEs averages, which are the value estimates of the next states.

\begin{figure*}[t]
\vskip 0.2in
\begin{center}
\centerline{
\includegraphics[width=0.7\columnwidth]{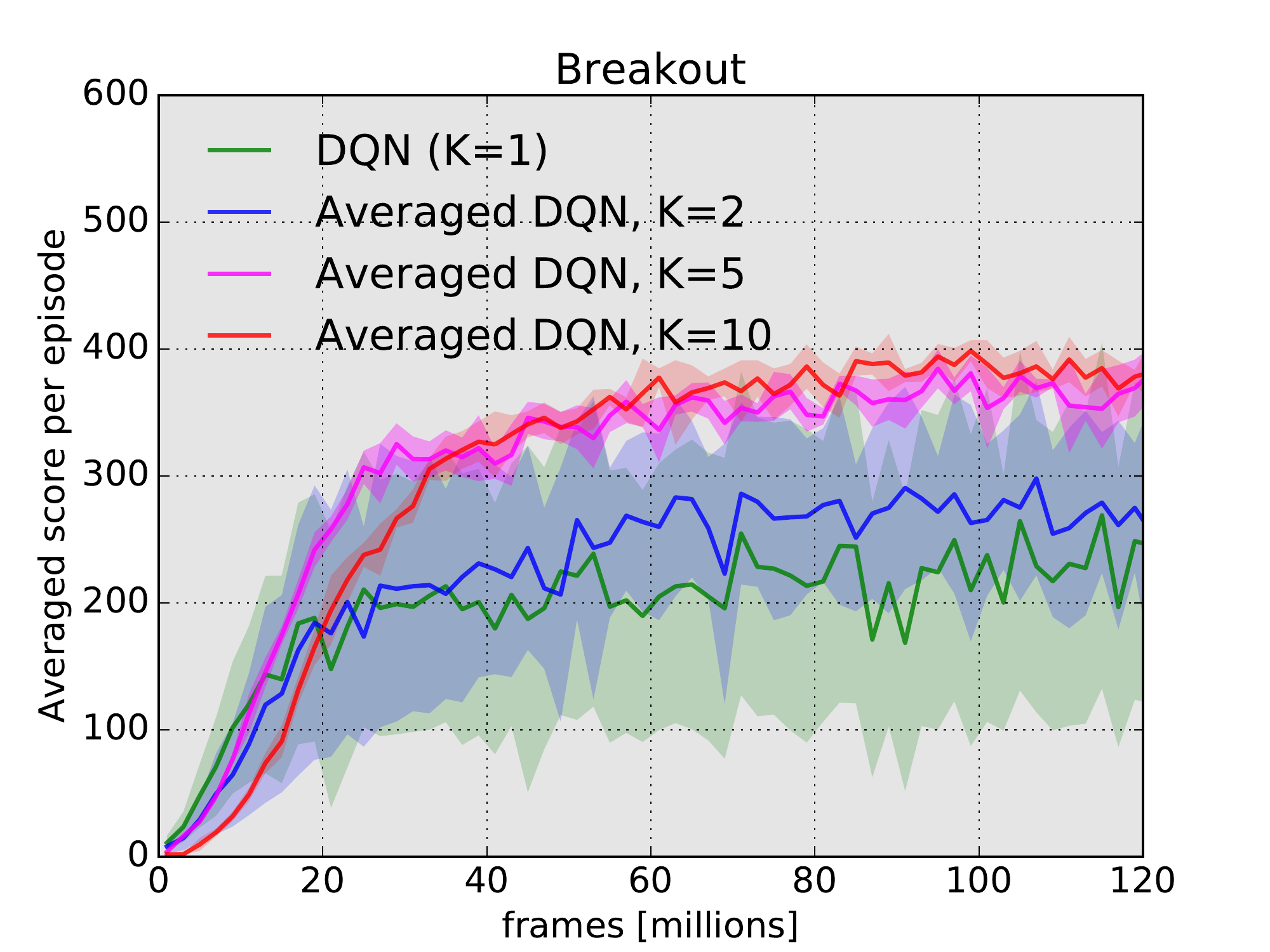}
\includegraphics[width=0.7\columnwidth]{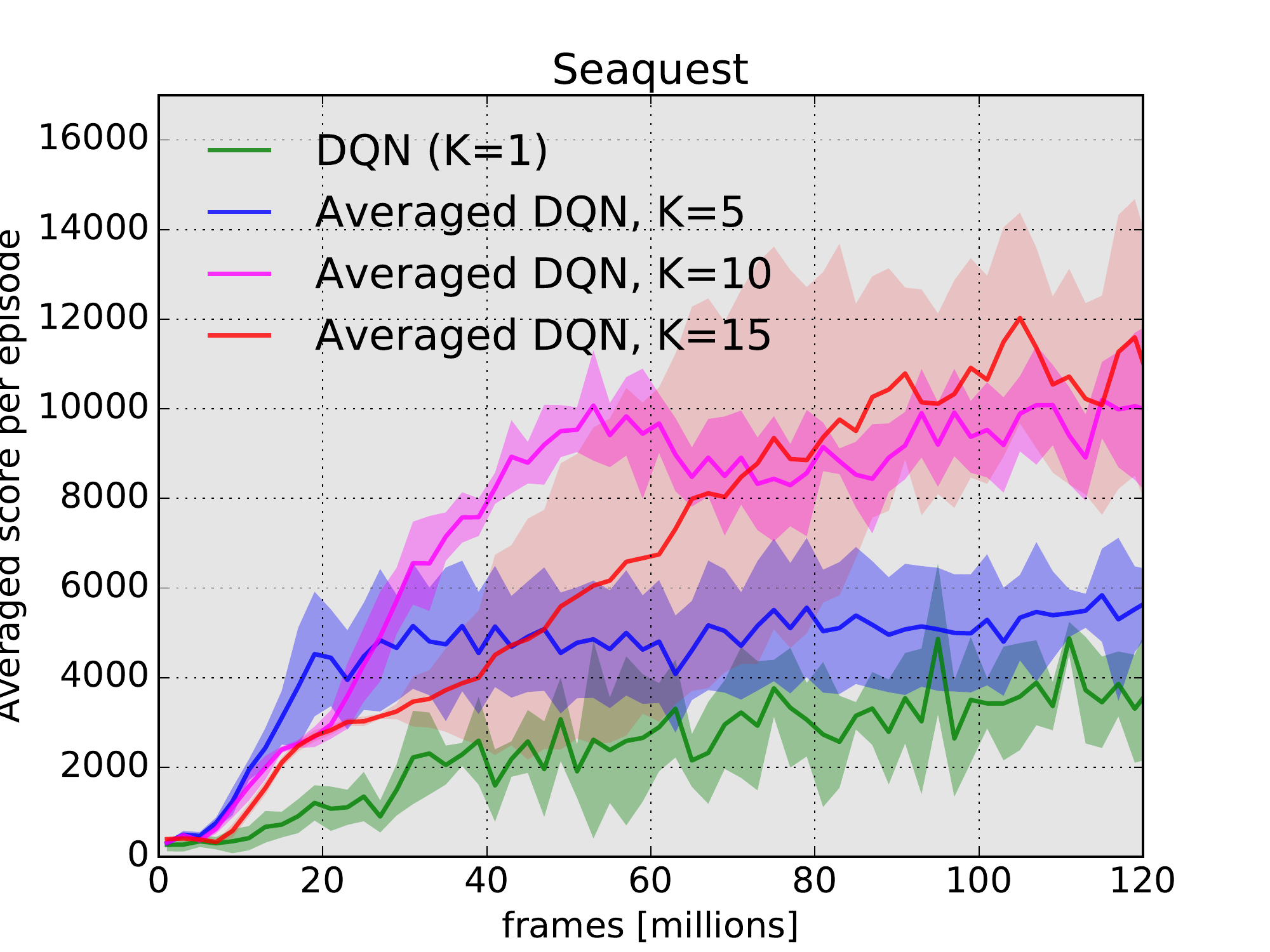}
\includegraphics[width=0.7\columnwidth]{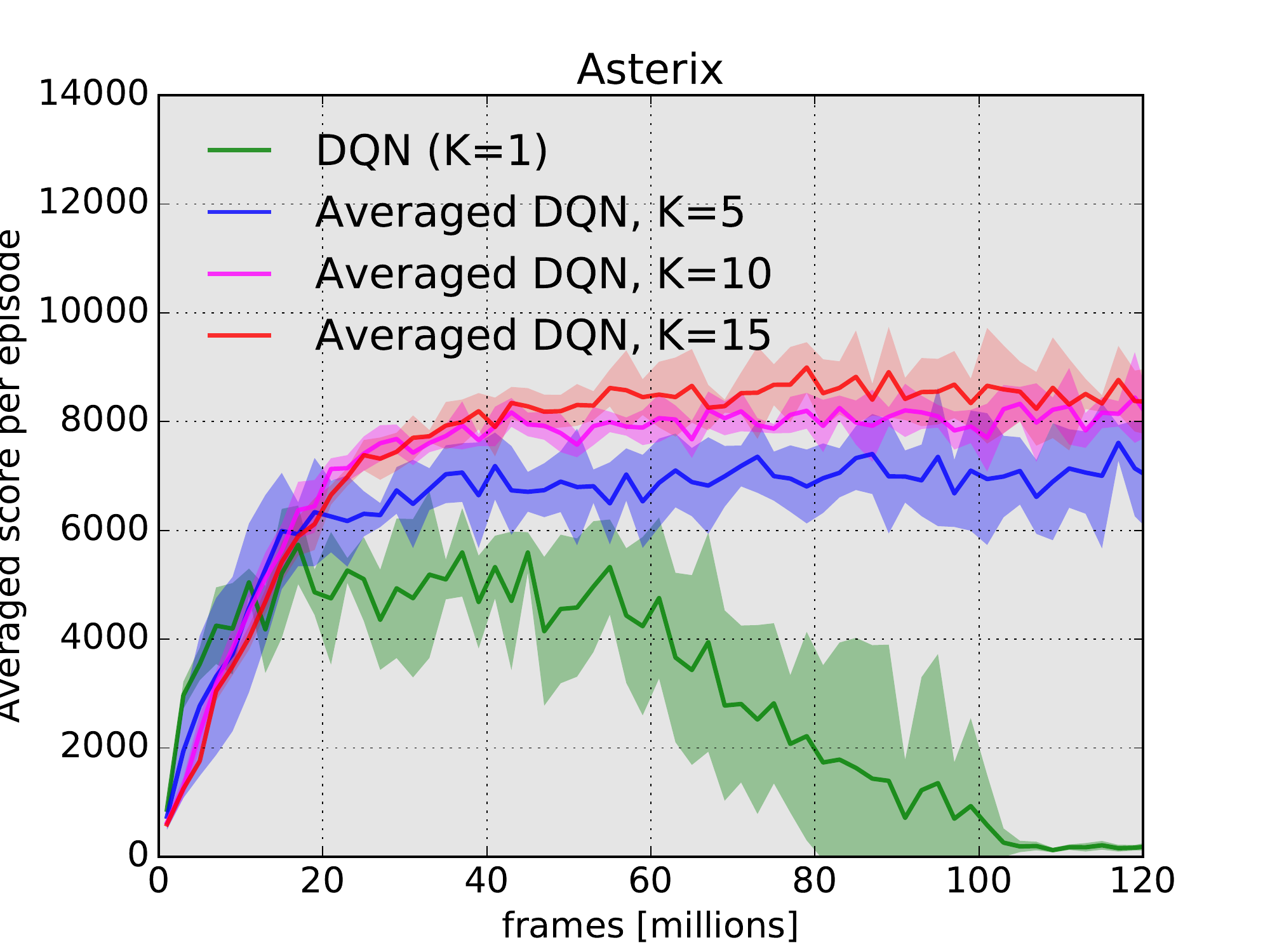}
}
\centerline{
\includegraphics[width=0.7\columnwidth]{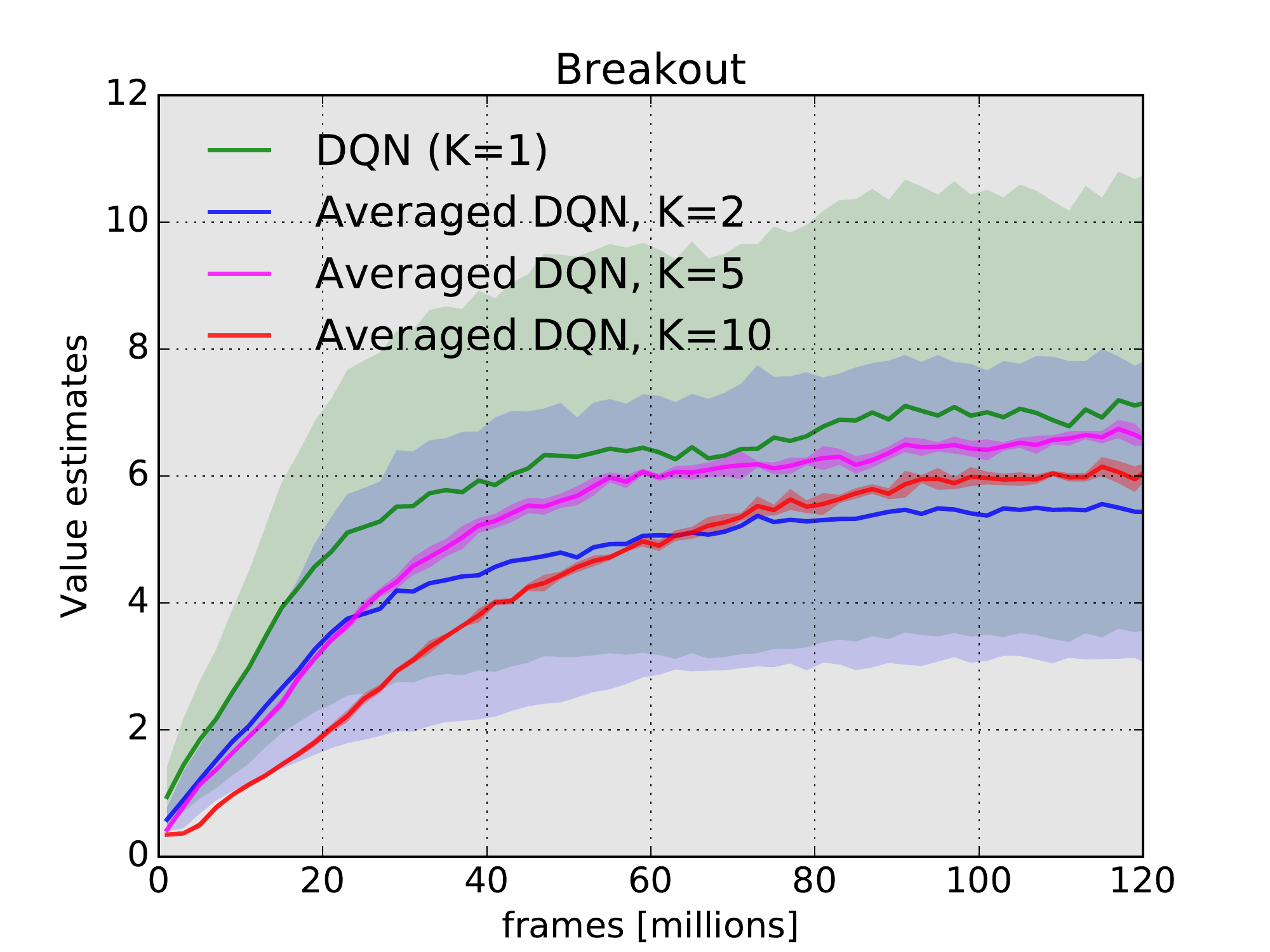}
\includegraphics[width=0.7\columnwidth]{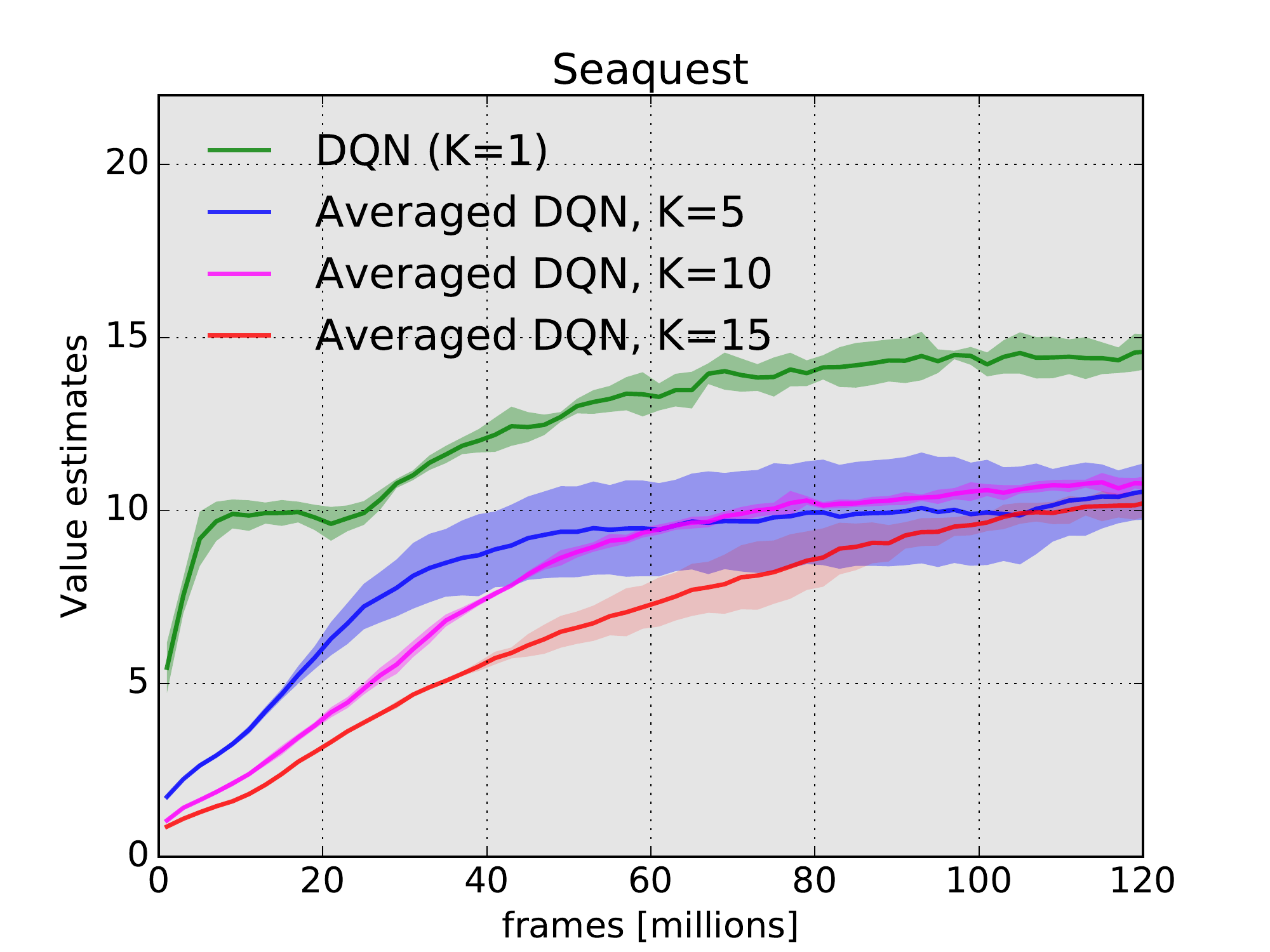}
\includegraphics[width=0.7\columnwidth]{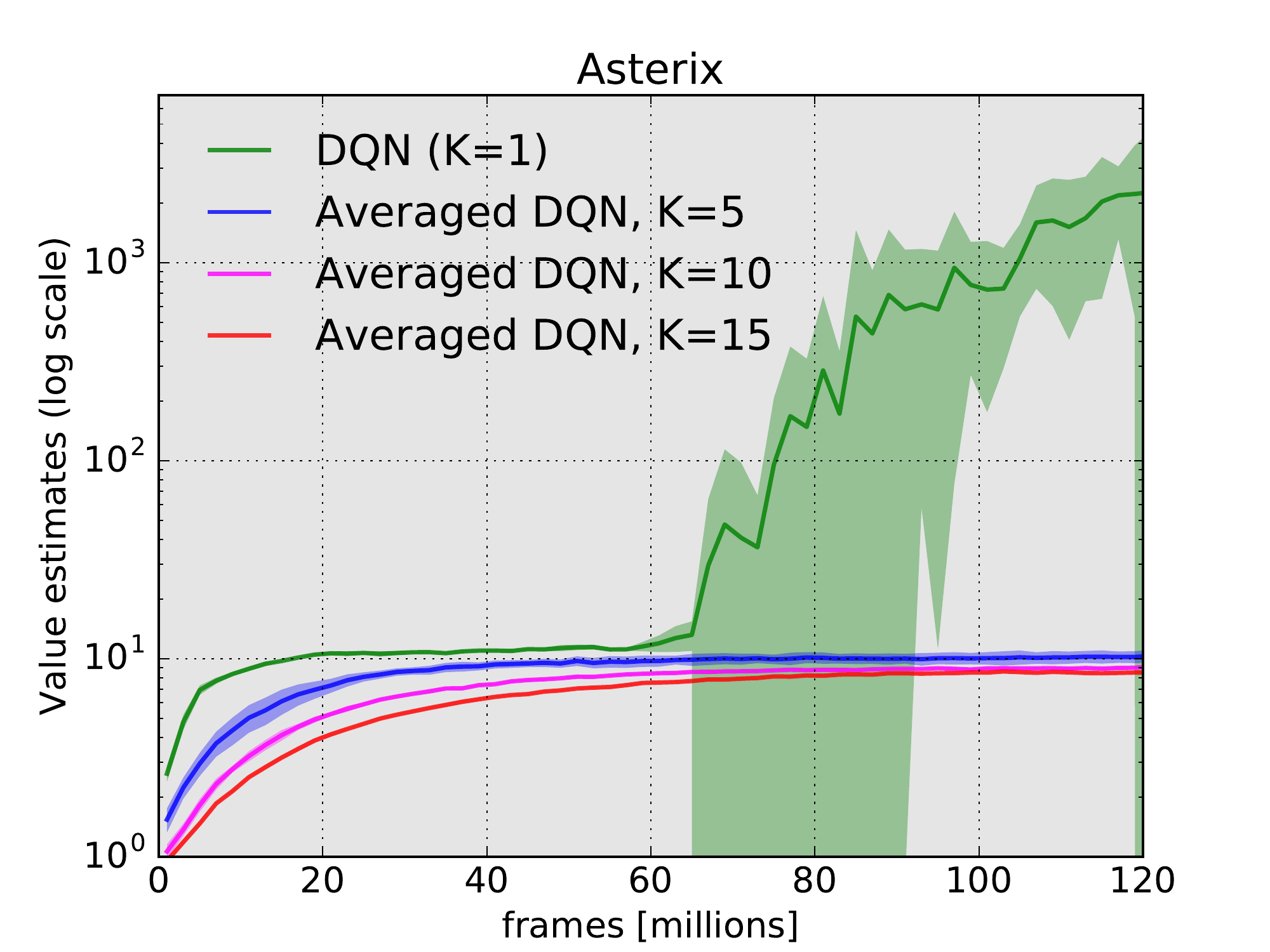}
}
\caption{The \textbf{top} row shows Averaged-DQN performance for the different number $K$ of averaged networks on three Atari games.
For $K=1$ Averaged-DQN is reduced to DQN.
The bold lines are averaged over seven independent learning trials.
Every 2M frames, a performance test using $\epsilon$-greedy policy with $\epsilon=0.05$ for 500000 frames was conducted.
The shaded area presents one standard deviation.
The \textbf{bottom} row shows the average value estimates for the three games.
It can be seen that as the number of averaged networks is increased, overestimation of the values is reduced, performance improves, and less variability is observed.
The hyperparameters used were taken from \citet{mnih2015human}.
}
\label{fig:experiments}
\end{center}
\vskip -0.2in
\end{figure*} 
\begin{table*}[t]
\caption{The columns present the average performance of DQN and Averaged-DQN after 120M frames, using $\epsilon$-greedy policy with $\epsilon=0.05$ for 500000 frames.
The standard variation represents the variability over seven independent trials.
Average performance improved with the number of averaged networks.
Human and random performance were taken from \citet{mnih2015human}.
}
\label{table:adqn_eperiments}
\vskip 0.15in
\begin{center}
\begin{small}
\begin{sc}
\begin{tabular}{r | c | c | c |c |c| r}
\hline
\abovespace\belowspace
Game & DQN  &  Averaged-DQN  & Averaged-DQN  & Averaged-DQN  & Human & Random \\
 & Avg. (std. dev.) &   (K=5) &  (K=10) &  (K=15) &  &  \\
\hline
\hline
\abovespace\belowspace
Breakout    & 245.1 \,\,\, (124.5)& 381.5  \,\,\, (20.2) &  381.8  \,\,\, (24.2) & - - & 31.8 & 1.7  \\
\hline
\abovespace\belowspace
Seaquest   & 3775.2 \,\,\, (1575.6) & 5740.2 \,\,\, (664.79
) &  9961.7 \,\,\, (1946.9) &  10475.1 \,\,\, (2926.6) & 20182.0& 68.4 \\
\hline
\abovespace\belowspace
Asterix     & 195.6 \,\,\, (80.4) & 6960.0  \,\,\, (999.2) & 8008.3 \,\,\, (243.6) & 8364.9 \,\,\, (618.6) & 8503.0 & 210.0 \\
\hline
\end{tabular}
\end{sc}
\end{small}
\end{center}
\vskip -0.1in
\end{table*}

\section{Experiments}
\label{sec:experiments}
The experiments were designed to address the following questions:
\begin{itemize}
\item How does the number $K$ of averaged target networks affect the error in value estimates, and in particular the overestimation error.
\item How does the averaging affect the learned polices quality.
%\item Does the quality of the policies improve as a result of averaging?
%\item Can the TAE and overestimation effects, discussed in Sections \ref{sec:oversetimation_and_approx} and \ref{sec:variance_reduction}, be observed in a small problem, where the optimal solution can be computed.
\end{itemize}
To that end, we ran Averaged-DQN and DQN on the ALE benchmark. Additionally, we ran  Averaged-DQN, Ensemble-DQN, and DQN on a Gridworld toy problem where the optimal value function can be computed exactly.

\subsection{Arcade Learning Environment (ALE)}
To evaluate Averaged-DQN, we adopt the typical RL methodology where agent performance is measured at the end of training.
We refer the reader to \citet{liang2016state} for further discussion about DQN evaluation methods on the ALE benchmark.
The hyperparameters used were taken from \citet{mnih2015human}, and are presented for completeness in Appendix \ref{appendix:dqn_hyper_params}.
DQN code was taken from McGill University RLLAB, and is available online\footnote{McGill University RLLAB DQN Atari code: \url{https://bitbucket.org/rllabmcgill/atari_release}.\\ Averaged-DQN code 
\url{https://bitbucket.org/oronanschel/atari_release_averaged_dqn}} (together with Averaged-DQN implementation).

%\url{https://bitbucket.org/oronanschel/atari_release_averaged_dqn}} (together with Averaged-DQN implementation).

We have evaluated the Averaged-DQN algorithm on three Atari games from the Arcade Learning Environment \citep{bellemare13arcade}.
The game of \textsc{Breakout} was selected due to its popularity and the relative ease of the DQN to reach a steady state policy.
In contrast, the game of \textsc{Seaquest} was selected due to its relative complexity, and the significant improvement in performance obtained by other DQN variants (e.g., \citet{schaul2015prioritized,dueling}).
Finally, the game of \textsc{Asterix} was presented in \citet{van2015deep} as an example to overestimation in DQN that leads to divergence.

As can be seen in Figure \ref{fig:experiments} and in Table~\ref{table:adqn_eperiments} for all three games, increasing the number of averaged networks in Averaged-DQN results in lower average values estimates, better-preforming policies, and less variability between the runs of independent learning trials.
For the game of \textsc{Asterix}, we see similarly to \citet{van2015deep} that the divergence of DQN can be prevented by averaging.

Overall, the results suggest that in practice Averaged-DQN reduces the TAE variance, which leads to smaller overestimation, stabilized learning curves and significantly improved performance.

\subsection{Gridworld}
The Gridworld problem (Figure~\ref{fig:gridworld}) is a common RL benchmark (e.g., \citet{boyan1995generalization}).
As opposed to the ALE, Gridworld has a smaller state space that allows the ER buffer to contain all possible state-action pairs.
Additionally, it allows the optimal value function $Q^*$ to be accurately computed.
\par
For the experiments, we have used Averaged-DQN, and Ensemble-DQN with ER buffer containing all possible state-action pairs.
The network architecture that was used composed of a small fully connected neural network with one hidden layer of 80 neurons.
For minimization of the DQN loss, the ADAM optimizer \citep{adam} was used on 100 mini-batches of 32 samples per target network parameters update in the first experiment, and 300 mini-batches in the second.

\subsubsection{Environment Setup}
\begin{figure}[ht]
\vskip 0.2in
\begin{center}
\centerline{
\includegraphics[width=0.40\columnwidth]{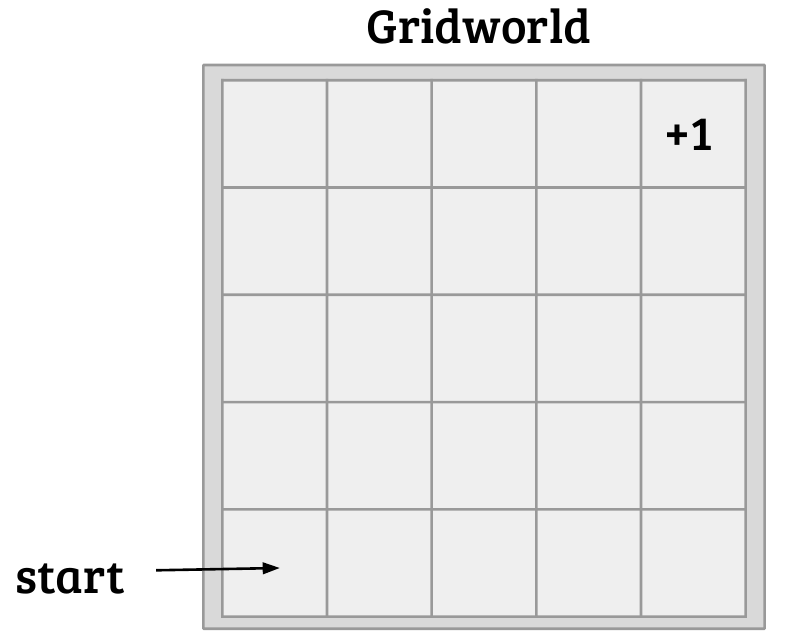}
}
\caption{Gridworld problem.
The agent starts at the left-bottom of the grid.
In the upper-right corner, a reward of +1 is obtained.
}
\label{fig:gridworld}
\end{center}
\vskip -0.2in
\end{figure}

In this experiment on the problem of Gridworld (Figure \ref{fig:gridworld}), the state space contains pairs of points from a 2D discrete grid ($S=\{(x,y)\}_{x,y\in 1,\ldots, 20}$). The algorithm interacts with the environment through raw pixel features with a one-hot feature map $ \phi(s_t):= (\mathds{1} \{s_t=(x,y)\})_{x,y\in 1,\ldots,20}$. There are four actions corresponding to steps in each compass direction, a reward of $r=+1$ in state $s_t=(20,20)$, and $r=0$ otherwise.
We consider the discounted return problem with a discount factor of $ \gamma = 0.9$.
\par

\subsubsection{Overestimation}
\begin{figure}[ht]
\vskip 0.2in
\begin{center}
\centerline{
\includegraphics[width=0.85\columnwidth]{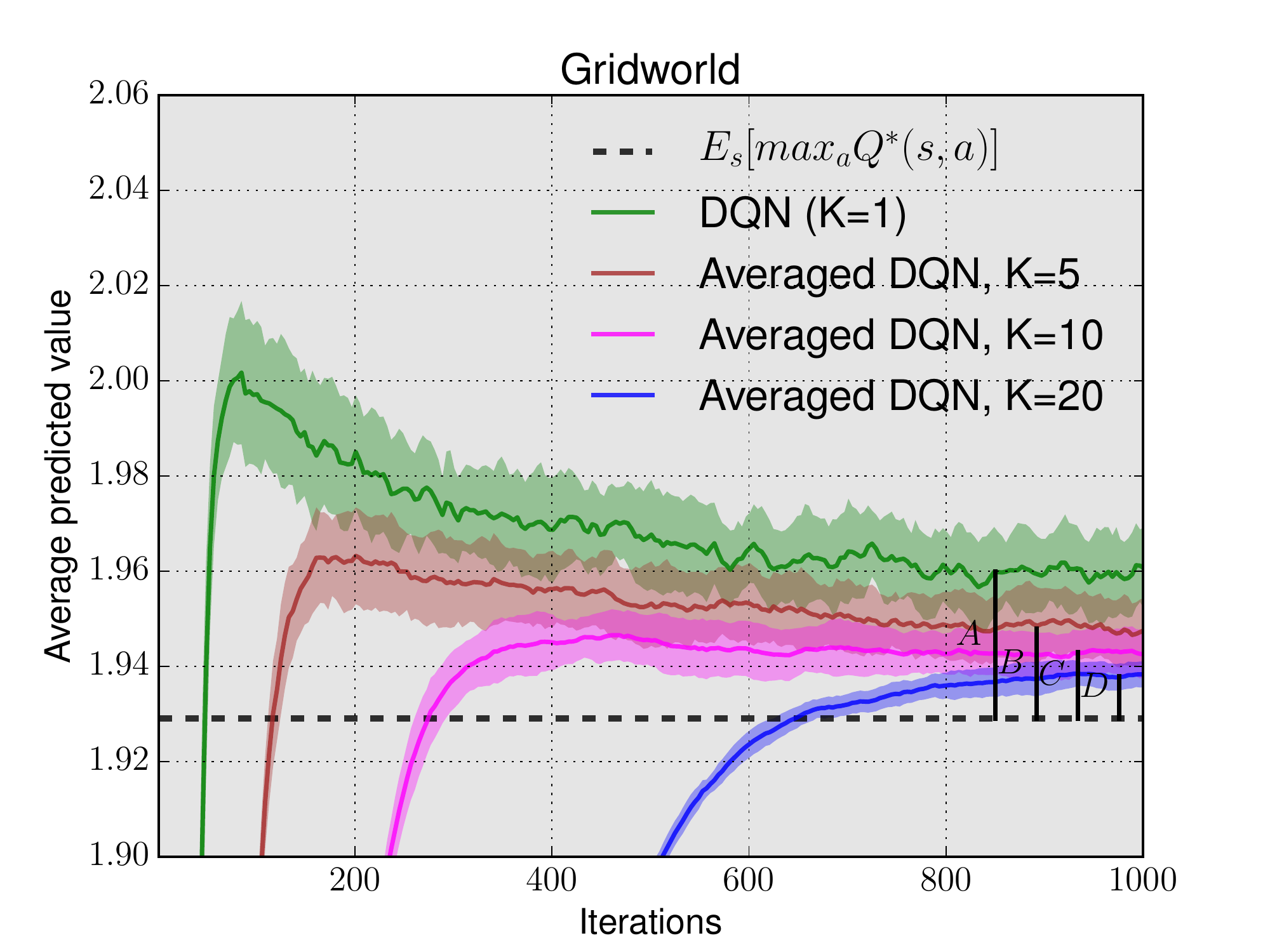}
}
\caption{Averaged-DQN average predicted value in Gridworld. Increasing the number $K$ of averaged target networks leads to a faster convergence with less overestimation (positive-bias).
The bold lines are averages over 40 independent learning trials, and the shaded area presents one standard deviation.
In the figure, \textbf{A,B,C,D} present DQN, and Averaged-DQN for K=5,10,20 average overestimation.  
}
\label{fig:gridworld_compare_k}
\end{center}
\vskip -0.2in
\end{figure}

In Figure \ref{fig:gridworld_compare_k} it can be seen that increasing the number $K$ of averaged target networks leads to reduced overestimation eventually.
Also, more averaged target networks seem to reduces the overshoot of the values, and leads to smoother and less inconsistent convergence. 
 
\subsubsection{Averaged versus Ensemble DQN}
\begin{figure}[ht]
\vskip 0.2in
\begin{center}
\centerline{
\includegraphics[width=0.85\columnwidth]{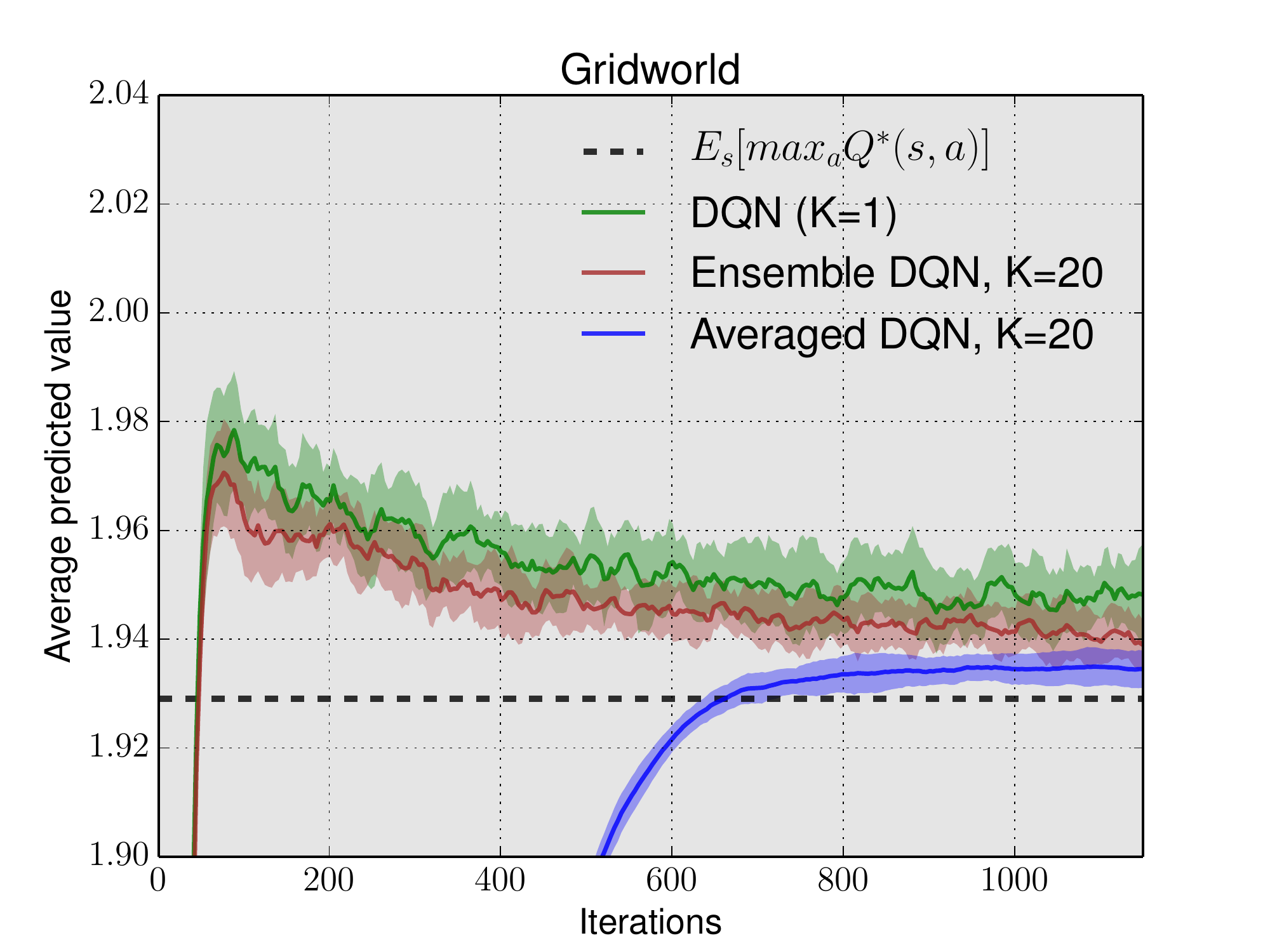}
}
\caption{Averaged-DQN and Ensemble-DQN predicted value in Gridworld. Averaging of past learned value is more beneficial than learning in parallel.
The bold lines are averages over 20 independent learning trials, where the shaded area presents one standard deviation.
}
\label{fig:gridworld_adqn_ensemble}
\end{center}
\vskip -0.2in
\end{figure}

In Figure \ref{fig:gridworld_adqn_ensemble}, it can be seen that as was predicted by the analysis in Section \ref{sec:variance_reduction}, Ensemble-DQN is also inferior to Averaged-DQN regarding variance reduction, and as a consequence far more overestimates the values.
We note that Ensemble-DQN was not implemented for the ALE experiments due to its demanding computational effort, and the empirical evidence that was already obtained in this simple Gridworld domain.

\section{Discussion and Future Directions}
In this work, we have presented the Averaged-DQN algorithm, an extension to DQN that stabilizes training, and improves performance by efficient TAE variance reduction.
We have shown both in theory and in practice that the proposed scheme is superior in TAE variance reduction, compared to a straightforward but computationally demanding approach such as Ensemble-DQN (Algorithm \ref{algo:Ensemble_DQN}).
We have demonstrated in several games of Atari that increasing the number $K$ of averaged target networks leads to better policies while reducing overestimation.
Averaged-DQN is a simple extension that can be easily integrated with other DQN variants such as \citet{schaul2015prioritized,van2015deep,dueling,bellemare2016unifying,playInADay}.
Indeed, it would be of interest to study the added value of averaging when combined with these variants.
Also, since Averaged-DQN has variance reduction effect on the learning curve, a more systematic comparison between the different variants can be facilitated as discussed in \cite{liang2016state}.

In future work, we may dynamically learn when and how many networks to average for best results.
One simple suggestion may be to correlate the number of networks with the state TD-error, similarly to \citet{schaul2015prioritized}.
%Finally, the Idea of introducing average over past learned value as an alternative to ensemble techniques could be implemented in other areas such as Generative Adverserial Networks.
Finally, incorporating averaging techniques similar to Averaged-DQN within on-policy methods such as SARSA and Actor-Critic methods \citep{mnih2016asynchronous} can further stabilize these algorithms.

\bibliography{mybib}
\bibliographystyle{icml2017}

\clearpage
\appendix

\section{DQN Variance Source Example}
\label{appendix:variance_between_models}
\begin{figure}[ht]
\vskip 0.2in
\begin{center}
\centerline{
\includegraphics[width=1\columnwidth]{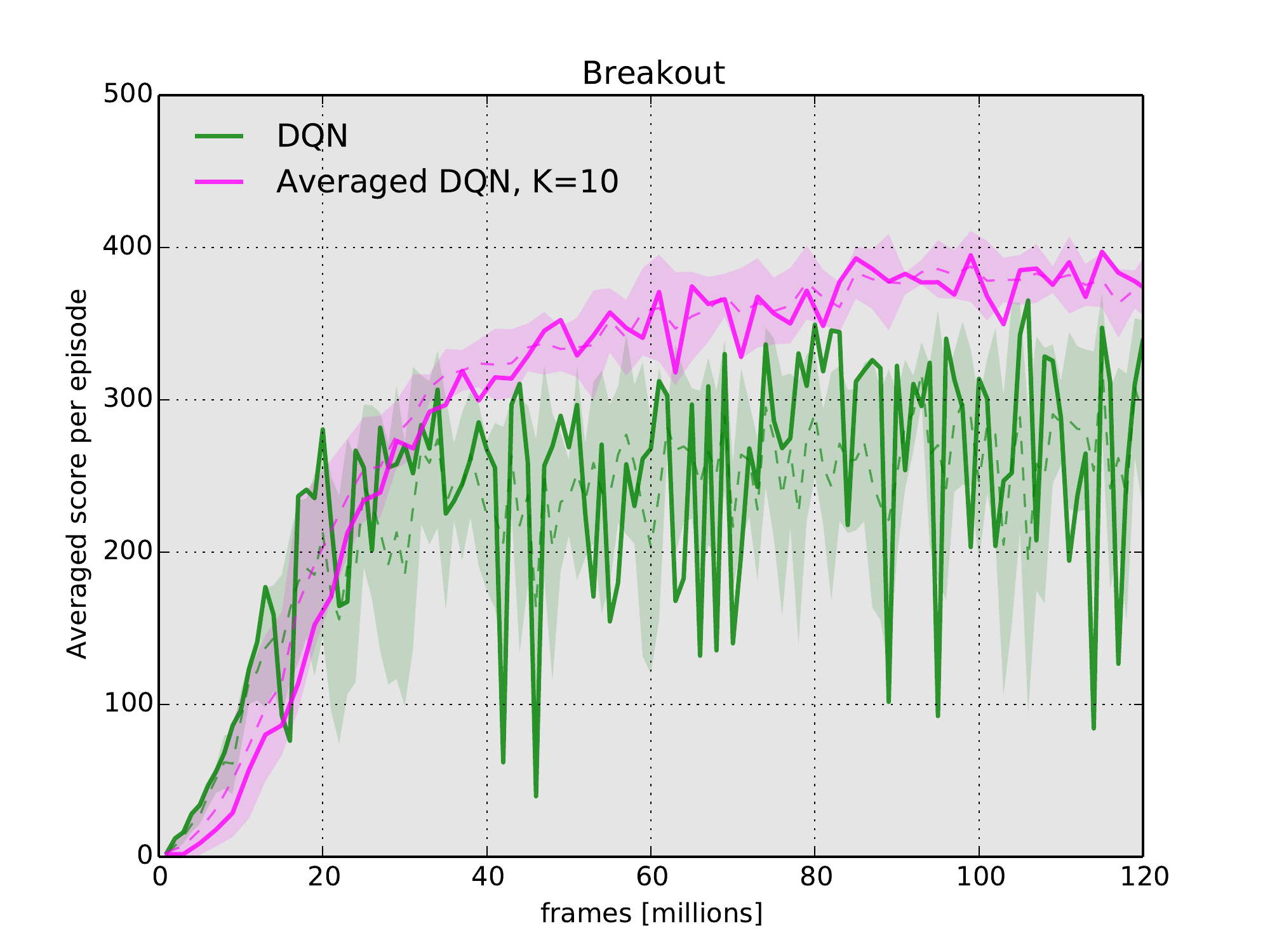}
}
\caption{DQN and Averaged-DQN performance in the Atari game of \textsc{Breakout}.
The\textbf{ bold lines} are \textbf{single learning trials} of the DQN and Averaged-DQN algorithm.
The dashed lines present average of 7 independent learning trials.
Every 1M frames, a performance test using $\epsilon$-greedy policy with $\epsilon=0.05$ for 500000 frames was conducted.
The shaded area presents one standard deviation (from the average).
For both DQN and Averaged-DQN the hyperparameters used, were taken from \citet{mnih2015human}.
}
\label{fig:vairance_between_models}
\end{center}
\vskip -0.2in
\end{figure}

Figure \ref{fig:vairance_between_models} presents a single learning trial of DQN compared to Averaged-DQN, which emphasizes that the source of variability in DQN between learning trials is due to occasions drops in average score within the learning trial.
As suggested in Section~\ref{sec:oversetimation_and_approx}, this effect can be related to the TAE causing to a deviation from the steady state policy.

%\section{DQN and Averaged DQN as Linear Filters}
%\label{appendix:general_analysis}
%\subsection{DQN}
%Let $V_i\in \mathbb{R}^{|S|}$ denote DQN value estimates for iteration $i$, let $P\in \mathbb{R}^{|S| \times |S|}_{+}$ denote the policy update   
%next state probabilities matrix, let $Z_i\in \mathbb{R}^{|S|}$ denote the TAEs vector we can write DQN value update equation as:
%\begin{align*}
%V_i = Z_i + \gamma PV_{i-1},
%\end{align*}
%taking the z-Transform and rearranging we get that,
%\begin{align*}
%\frac{V(z)}{Z(z)} = [I-\gamma Pz^{-1}]^{-1}
%\end{align*}

\section{Ensemble DQN TAE Variance Calculation in a unidirectional MDP (Section~\ref{sec:Ensemble DQN Variance})}
\label{appendix:ensemble_dqn_two_state_cal}
Recall that $\mathbb{E}[Z^{k,i}_{s,a}]=0$, $\text{Var}[Z^{k,i}_{s,a}] = \sigma^2_s $, for all $i \neq j$: $\text{Cov}[Z^{k,i}_{s,a},Z^{k',j}_{s',a}]=0$, and for all $k \neq k'$: $\text{Cov}[Z^{k,i}_{s,a},Z^{k',j}_{s',a}]=0$.
Following the Ensemble-DQN update equations in \Algo{\ref{algo:Ensemble_DQN}}:

\begin{align*}
& Q^{E}_i(s_0,a) =\\
& = \frac{1}{K}\sum^{K}_{k=1} Q(s_0,a;\theta^{k}_{i})\\
& = \frac{1}{K}\sum^{K}_{k=1}[Z^{k,i}_{s_0,a} + y^{i}_{s_0,a}]\\
& = \frac{1}{K}\sum^{K}_{k=1}[Z^{k,i}_{s_0,a}] + y^{i}_{s_0,a}\\
& = \frac{1}{K}\sum^{K}_{k=1}[Z^{k,i}_{s_0,a}] + \gamma Q^{E}_{i-1}(s_1,a)\\
& = \frac{1}{K}\sum^{K}_{k=1}[Z^{k,i}_{s_0,a}] +  \frac{\gamma}{K}\sum^{K}_{k=1}[Z^{k,i-1}_{s_1,a}] + \gamma y^{i-1}_{s_2,a}.
\end{align*}
By iteratively expanding $y^{i-1}_{s_2,a}$ as above, and noting that $y^{j}_{s_{M-1},a}=0$ for all times (terminal state), we obtain,
\begin{align*}
Q^{E}_i(s_0,a) = \sum^{M-1}_{m=0} \gamma^m  \frac{1}{K}\sum^{K}_{k=1} Z^{k,i-m}_{s_m,a}.
\end{align*}
Since the TAEs are uncorrelated by assumption, we get
\begin{align*}
\text{Var}[Q^{E}_i(s_0,a)] = \sum_{m=0}^{M-1}  \frac{1}{K} \gamma^{2m} \sigma^2_{s_m}.
\end{align*}

\section{Averaged DQN TAE Variance Calculation in a unidirectional MDP (Section~\ref{sec:Averaged DQN Variance})}
\label{appendix:adqn_two_state_cal}
Recall that $\mathbb{E}[\ApproxVar]=0$, $\text{Var}[\ApproxVar] = \sigma^2_s $, and for $i \neq j$: $\text{Cov}[\ApproxVar,Z^{j}_{s',a'}]=0$.
Further assume that for all $s \neq s'$: $\text{Cov}[\ApproxVar,Z^{i}_{s',a}]=0$.
Following the Averaged-DQN update equations in \Algo{\ref{algo:ADQN}}:
\begin{align*}
& Q^{A}_i(s_0,a) =\\
& = \frac{1}{K}\sum^{K}_{k=1} Q(s_0,a;\theta_{i+1-k})\\
& = \frac{1}{K}\sum^{K}_{k=1}[Z^{i+1-k}_{s_0,a} + y^{i+1-k}_{s_0,a}]\\
& = \frac{1}{K}\sum^{K}_{k=1}[Z^{i+1-k}_{s_0,a}] + \frac{\gamma}{K}\sum^{K}_{k=1}Q^{A}_{i-k}(s_1,a)\\
& = \frac{1}{K}\sum^{K}_{k=1}[Z^{i+1-k}_{s_0,a}] + \frac{\gamma}{K^2}\sum^{K}_{k=1}\sum^{K}_{k'=1}Q(s_1,a;\theta_{i+1-k-k'}).\\
\end{align*}
By iteratively expanding $Q(s_1,a;\theta_{i+1-k-k'})$ as above, and noting that $y^{j}_{s_{M-1},a}=0$ for all times (terminal state), we get
\begin{align*}
& Q^{A}_i(s_0,a) = \frac{1}{K}\sum^{K}_{k=1}Z^{i+1-k}_{s_0,a}  + \frac{\gamma}{K^2}\sum^{K}_{k=1}\sum^{K}_{k'=1}Z^{i+1-k-k'}_{s_{1},a}\\
&+ \cdots + \frac{\gamma^{M-1}}{K^M}\sum^{K}_{j_1=1}\sum^{K}_{j_2=1}\cdots \sum^{K}_{j_{M}=1} Z^{i+1-j_1-\cdots -j_M}_{s_{M-1},a} .
\end{align*}

Since the TAEs in different states are uncorrelated by assumption the latter sums are uncorrelated and we may calculate the variance of each separately. For $L=1,\ldots,M$, denote
\begin{align*}
V_L=\text{Var} \left[ \frac{1}{K^L}\sum^{K}_{i_1=1} \sum^{K}_{i_1=2} \ldots \sum^{K}_{i_L=1} Z_{ i_1 + i_2 + \ldots + i_L}  \right],
\end{align*}
where to simplify notation $Z_{L},Z_{L+1},\ldots,Z_{K\cdot L}$ are independent and identically distributed (i.i.d.\ ) TAEs random variables, with $\mathbb{E}[Z_l]=0$ and $\mathbb{E}[(Z_l)^2]=\sigma^2_z$.

Since the random variables are zero-mean and i.i.d.\ we have that:
\begin{align*}
 V_L &=\frac{1}{K^{2L}} \mathbb{E}_z \left[ \right( \sum^{KL}_{j=L} n_{j,L} Z_{j} \left)^2 \right]\\
 &= \frac{\sigma^2_z}{K^{2L}}  \sum^{KL}_{j=L}  (n_{j,L})^2 ,
\end{align*}

where $n_{j,L}$ is the number of times $Z_j$ is counted in the multiple summation above.
The problem is reduced now to evaluating $n_{j,L}$ for $j\in \{ L,\ldots, K\cdot L \}$, where $n_{j,L}$ is the number of solutions for the following equation:
\begin{equation}
\label{eq:dual_problem_to_DKM}
i_1+i_2+\ldots+i_L=j,
\end{equation}
over $i_1,\ldots,i_L \in \{1,\ldots,K\}$. The calculation can be done recursively, by noting that
\begin{align*}
n_{j,L} = \sum^{K}_{i=1} n_{j-i,L-1}.
\end{align*}

Since the final goal of this calculation is bounding the variance reduction coefficient, we will calculate the solution in the frequency domain where the bound can be easily obtained.

Denote
\[ u^K_j =
\begin{cases}
1 \,\, \text{if}  \,\, j \in \{1,\ldots,K\}\\
0 \,\, \text{otherwise} \qquad\qquad,
\end{cases}
\]
For $L=1$ (base case), we trivially get that for any $j\in \mathbb{Z}$: 
\begin{align*}
n_{j,1}= u^K_j,
\end{align*}
and we can rewrite our recursive formula for $n_{j,L}$ as:
\begin{align*}
n_{j,L} &= \sum^{\infty}_{i=-\infty} n_{j-i,L-1} \cdot u^K_i\\
&\equiv (n_{j-1,L-1} * u^K)_j\\
&=\underset{L \,\,\text{times}}{\underbrace{(u^K*u^K \ldots *u^K})_j},
\end{align*}
where $*$ is the discrete convolution.

To continue, we denote the Discrete Fourier Transform (DFT) of $u^K=(u_n^K)_{n=0}^{N-1}$ as $U=(U_n)_{n=0}^{N-1}$, and by using Parseval's theorem we have that
\begin{align*}
& V_L= 
\frac{\sigma^2_z }{K^{2L}} \sum^{N-1}_{n=0} |(u^K * u^K\ldots *u^K)_n|^2\\
&=\frac{\sigma^2_z }{K^{2L}}\frac{1}{N}\sum^{N-1}_{n=0}|U_n|^{2L},
\end{align*} 
where $N$ is the length of the vectors $u$ and $U$ and is taken large enough so that the sum includes all non-zero elements of the convolution. We denote $D_{K,m} = \frac{1}{K^{2(m+1)}}\frac{1}{N}\sum^{N-1}_{n=0}|U_n|^{2(m+1)}$, and now we can write Averaged-DQN variance as:
\begin{align*}
\text{Var}[Q^{A}_i(s_0,a)] &= \sum_{m=0}^{M-1} D_{K,m} \gamma^{2m} \sigma^2_{s_m}.
\end{align*}

Next we bound $D_{K,m}$ in order to compare Averaged-DQN to Ensemble-DQN, and to DQN.

For $K>1,m>0$:
\begin{align*}
D_{K,m} &= \frac{1}{K^{2(m+1)}} \frac{1}{N} \sum^{N-1}_{n=0}|U_n|^{2(m+1)}\\
&=  \frac{1}{N} \sum^{N-1}_{n=0}|U_n/K|^{2(m+1)}\\
&<  \frac{1}{N} \sum^{N-1}_{n=0}|U_n/K|^2\\
&=  \frac{1}{K^2} \sum^{N-1}_{n=0}|u^K_n|^2\\
&= \frac{1}{K}
\end{align*}

where we have used the easily verified facts that $\frac{1}{K}|U_n|\leq 1$, $\frac{1}{K}|U_n|= 1$ only if $n=0$, and Parseval's theorem again.

\section{Experimental Details for the Arcade Learning Environment Domain}
\label{appendix:dqn_hyper_params}
We have selected three popular games of Atari to experiment with Averaged-DQN.
We have used the exact setup proposed by \citet{mnih2015human} for the experiments, we only provide the details here for completeness.
The full implementation is available at \url{https://bitbucket.org/oronanschel/atari_release_averaged_dqn}.
%\url{https://bitbucket.org/oronanschel/atari_release_averaged_dqn}.

Each episode starts by executing a no-op action for one up to 30 times uniformly.
We have used a frame skipping where each agent action is repeated four times before the next frame is observed.
The rewards obtained from the environment are clipped between -1 and 1.
\subsection{Network Architecture}
The network input is a 84x84x4 tensor.
It contains a concatenation of the last four observed frames.
Each frame is rescaled (to a 84x84 image), and gray-scale.
We have used three convolutional layers followed by a fully-connected hidden layer of 512 units.
The first convolution layer convolves the input with 32 filters of size 8 (stride 4), the second, has 64 layers of size 4 (stride 2), the final one has 64 filters of size 3 (stride 1).
The activation unit for all of the layers a was Rectifier Linear Units (ReLu). 
The fully connected layer output is the different Q-values (one for each action).
For minimization of the DQN loss function RMSProp (with momentum parameter 0.95) was used.
\subsection{Hyper-parameters}
The discount factor was set to $\gamma$ = 0.99, and the optimizer learning rate to $\alpha$ = 0.00025.
The steps between target network updates were 10,000. Training is done over 120M frames. 
The agent is evaluated every 1M/2M steps (according to the figure).
The size of the experience replay memory is 1M tuples.
The ER buffer gets sampled to update the network every 4 steps with mini batches of size 32.
The exploration policy used is an $\epsilon$-greedy policy with $\epsilon$ decreasing linearly from 1 to
0.1 over 1M steps.

\end{document}